\documentclass[letterpaper, 10 pt, conference]{ieeeconf}  % Comment this line out
\IEEEoverridecommandlockouts                              % This command is only
\usepackage{xcolor}
\usepackage{comment}
\usepackage{graphics} % for pdf, bitmapped graphics files
\usepackage{graphicx}
\usepackage{subcaption}
\usepackage{epsfig} % for postscript graphics files
\usepackage{amsmath} % assumes amsmath package installed
\usepackage{amssymb}  % assumes amsmath package installed
\usepackage{mathtools}
\usepackage{marginnote}
\usepackage{bm}
\usepackage{hyperref}
\usepackage{iitem}
\usepackage{float}
\usepackage{algorithm}
\usepackage{algpseudocode}
\usepackage{booktabs}
\usepackage{svg}
\usepackage{balance}
\usepackage{caption}
\title{\LARGE \bf
\textsc{LOAD}: Locomotion with Online Awareness of Dynamics for \\Precision Footholds} % may change the name 

\author{
Kunzhao Ren$^{1}$, Zhishuo Zhu$^{2}$, and Xiaobin Xiong$^{1,\dagger}$%
\thanks{$^{1}$Shanghai Innovation Institute.}%
\thanks{$^{2}$University of Wisconsin-Madison.}%
\thanks{$^{\dagger}$Corresponding author: \texttt{xiaobin.xiong@sii.edu.cn}.}%
}

\usepackage{xcolor}
\usepackage{etoolbox}

\begin{document}

 \newcommand{\block}[1]{\noindent{\textbf{#1}:}}
\newcommand{\fix}[1]{{\color{red} #1}}

\newcommand{\todoRen}[1]{{\color{blue} #1}}

\newcommand{\todo}[1]{{\color{red} #1}}

\newcommand{\policy}[1]{\textit{#1}}

\maketitle
\thispagestyle{empty}
\pagestyle{empty}

\begin{abstract}
Unknown payloads alter robot dynamics and make precise contact
execution difficult when foothold margins are small. We present
\textbf{LOAD}, a perceptive locomotion framework that combines explicit
inertial estimates with implicit dynamics features for narrow-beam
traversal. Separate recurrent modules infer torso inertial changes
and task-relevant history from proprioception, while a height-map
teacher is distilled into a depth-based student. We extensively evaluate LOAD and its variants in simulation and on hardware across beam widths and payload configurations. The results show that implicit dynamics features accelerate training, while explicit inertial estimation further improves traversal success; combining both enables LOAD to approach the performance of a privileged policy supplied with ground-truth inertial parameters. Hardware experiments demonstrate traversal with unilateral payloads of
6\,kg on a 10\,cm beam and 9\,kg on a 15\,cm beam, and an evenly
distributed 11\,kg payload on a 15\,cm beam. These results support
combining physical estimates and history features for precision
locomotion under unknown loads.
\end{abstract}

\section{Introduction}
\label{sec:introduction}

Legged robots have demonstrated increasingly agile locomotion over rough and unstructured environments. Learning-based controllers have enabled quadrupeds to traverse challenging terrain with dynamic and robust motions~\cite{hwangbo2019learning,lee2020learning,rudin2022learning}. The integration of exteroceptive sensing has further improved locomotion over stairs, gaps, obstacles, and parkour-like environments~\cite{miki2022learning,cheng2024extreme,hoeller2024anymal}. Recent studies have also moved toward sparse and risky footholds through terrain-aware learning, terrain reconstruction, and foothold-aware perception~\cite{zhang2024risky,wang2025beamdojo, yu2024terrain,start2026,wang2026puma}. Complementary work incorporates explicit constraints into locomotion learning~\cite{kim2024constraints}. These developments highlight the importance of determining where and how each contact should be established.

Perceptive locomotion methods emphasize learning terrain representations to guide contact placement. Robustness to variations in robot dynamics is commonly promoted through domain randomization~\cite{peng2018sim,tan2018sim}, while Rapid Motor Adaptation (RMA) and related history-based estimation methods use proprioceptive history to adapt to changing conditions~\cite{kumar2021rma,ji2022concurrent,nahrendra2023dreamwaq}. Building on these advances, we investigate how implicit dynamics features and explicit inertial estimates jointly support precise contact execution on narrow beams under unknown payloads.

Existing narrow-beam studies report real-world traversal of an 18\,cm-wide single-plank bridge with Go2~\cite{dong2025marg} and 20\,cm-wide beams separated by gaps with ANYmal-D~\cite{zhang2024risky}. An A1 equipped with a reaction-wheel module has also traversed a beam less than 6\,cm-wide with external motion capture~\cite{lee2023enhanced}. More recently, START demonstrated traversal of 20\,cm-wide indoor balance beams and 15\,cm-wide outdoor flowerbed edges with Lite3 using onboard depth sensing and terrain reconstruction~\cite{start2026}. Complementary progress has been made in payload-adaptive locomotion. Rapid Embodiment Adaptation infers physical parameters from short interaction histories and conditions a locomotion policy on these estimates, enabling Go2 to adapt to a 5\,kg payload~\cite{li2026rapid}. FLAP combines physics-informed dynamics prediction with adaptive control to compensate for disturbances from shifting payloads, demonstrating Go1 locomotion with 5\,L of water across varied outdoor terrains~\cite{youngquist2026flap}. These results motivate studying payload adaptation under the restricted support geometry of narrow beams. Unknown payloads change mass, center-of-mass location, and rotational inertia, so the same contact forces can produce different body motions. The beam simultaneously limits lateral foot placement and corrective stepping, leaving little room to compensate for these changes. Reliable traversal therefore requires adapting contact behavior and body motion to the current dynamics within the available support geometry.

\begin{figure}[!t]
    \centering
    \includegraphics[width=\linewidth]
    {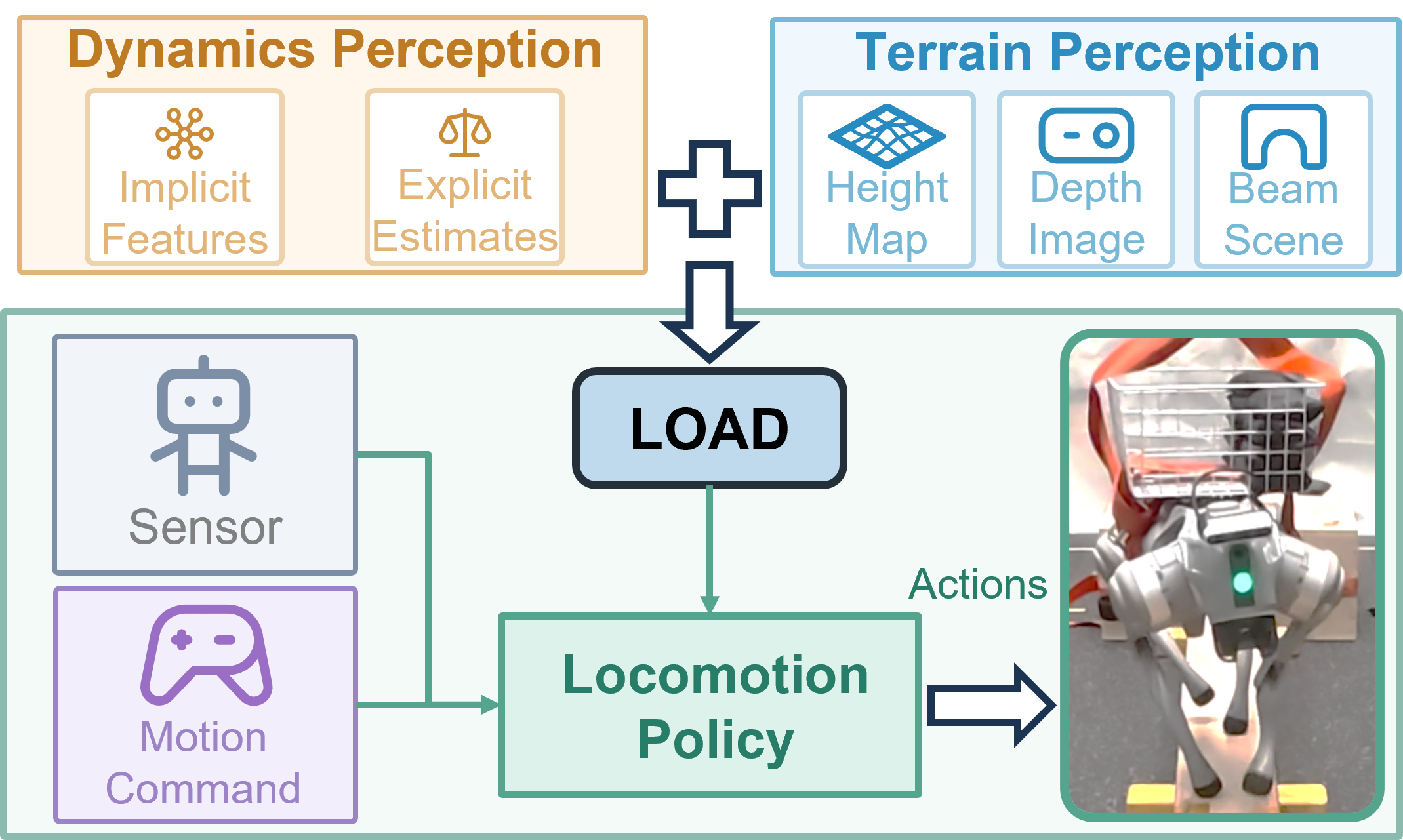}
   \caption{Overview of LOAD. Terrain perception, implicit dynamics features, and explicit inertial estimates jointly guide narrow-beam traversal under unknown payloads.}
    \label{fig:overview}
\end{figure}

To address this challenge, we present LOAD, a locomotion framework that integrates terrain perception with implicit dynamics features and explicit inertial estimates for narrow-beam traversal under unknown payloads. Two separate branches process proprioceptive history: one learns task-relevant dynamics features through reinforcement learning, while the other is supervised to estimate payload-induced changes in the torso's inertial properties relative to a nominal model. This pairing provides physically interpretable information about the current dynamics while retaining task-relevant history that does not need to be expressed through inertial parameters. The policy uses both outputs together with terrain features to adapt whole-body motion and foothold execution. A teacher trained with privileged height maps is distilled into a student using onboard depth images for deployment.

We evaluate LOAD in simulation and on hardware across beam widths and payload configurations. In the reported training runs, policies with implicit dynamics features reach high traversal success earlier than their counterparts without these features. These features also improve evaluated traversal performance, with benefits persisting even when ground-truth inertial parameters are available. Combining these features with explicit inertial estimates brings the LOAD teacher closer to its privileged counterpart, LOAD-GT, over the tested conditions. The benefit of combining explicit and implicit dynamics information persists after visual distillation on narrow beams within the training payload range. Tests with heavier, unseen payloads reveal limitations in both inertial identification and policy generalization, indicating that accurate dynamics information alone does not ensure successful extrapolation.

The contributions of this work are threefold:

\begin{itemize}
\item We develop LOAD, a perceptive locomotion framework that combines
online inertial estimation with implicit dynamics features to adapt
whole-body motion and foothold execution to unknown torso payloads.

\item We evaluate six policy variants to characterize the separate and combined benefits of explicit inertial information and implicit history. The comparisons show improved narrow-beam traversal with the combined representation and identify limitations beyond the training payload range.

\item We demonstrate narrow-beam traversal on a Unitree Go2 with
unknown unilateral payloads of 6\,kg on a 10\,cm beam and 9\,kg on
a 15\,cm beam, as well as an evenly distributed 11\,kg payload on
a 15\,cm beam.
\end{itemize}

\section{Related Work}
\label{sec:related_work}

\subsection{Perceptive and Precision Locomotion}

Proprioceptive reinforcement learning has enabled robust quadrupedal
locomotion over challenging terrain
~\cite{hwangbo2019learning,lee2020learning}.
Perceptive approaches incorporate elevation maps or depth observations
to anticipate terrain geometry~\cite{rudin2022learning,miki2022learning,cheng2024extreme,hoeller2024anymal}.
For sparse footholds, recent methods use terrain-aware learning~\cite{zhang2024risky,wang2025beamdojo}, terrain reconstruction~\cite{yu2024terrain,start2026}, and foothold priors~\cite{wang2026puma}. Complementary work incorporates explicit constraints into policy optimization to reduce reward engineering and guide locomotion behavior~\cite{kim2024constraints}. Model-based approaches address contact
feasibility through centroidal planning, trajectory optimization,
and multi-contact optimal control~\cite{dai2014whole,winkler2018gait,mastalli2020crocoddyl}.
These studies motivate our focus on how payload-induced dynamics
changes affect the execution of contacts on narrow support surfaces.

\subsection{Robust and Adaptive Locomotion}

Dynamics randomization improves robustness by exposing policies to
variations in physical parameters during training
~\cite{peng2018sim,tan2018sim}. RMA infers a latent dynamics context from recent observations and actions to condition a policy trained with privileged information~\cite{kumar2021rma}. Concurrent estimation
and control~\cite{ji2022concurrent} and learned terrain and state
representations~\cite{nahrendra2023dreamwaq} further demonstrate the
value of history-based inference. Such representations can retain
control-relevant information without explicitly identifying inertial
parameters. LOAD preserves this implicit pathway while adding
supervised inertial estimates. Our comparisons examine the separate
and combined contributions of these two information sources.

\subsection{Physical Parameter Estimation}

System identification and adaptive control use observed responses
to recover uncertain physical parameters~\cite{slotine1991applied,sastry1989adaptive}. For legged robots,
kinematics and contact constraints support state estimation~\cite{bloesch2012state,hartley2020contact}, while constrained
moving-horizon estimation jointly recovers states and ground-reaction
forces~\cite{kang2025mhe}. SPI-Active combines sampling-based
parameter identification with active exploration for sim-to-real
transfer~\cite{sobanbabu2025spi}. PRIME jointly estimates motion,
contact forces, and inertial parameters through physically consistent
optimization~\cite{prime2026}. LOAD uses a learned recurrent
identifier to supply online inertial estimates directly to the
locomotion policy. Its evaluation centers on how these estimates
complement implicit history in precision foothold control.

\section{LOAD: Locomotion with Online Awareness of Dynamics}
\label{sec:method}

\begin{figure}[t]
    \centering
    \includegraphics[width=\columnwidth]{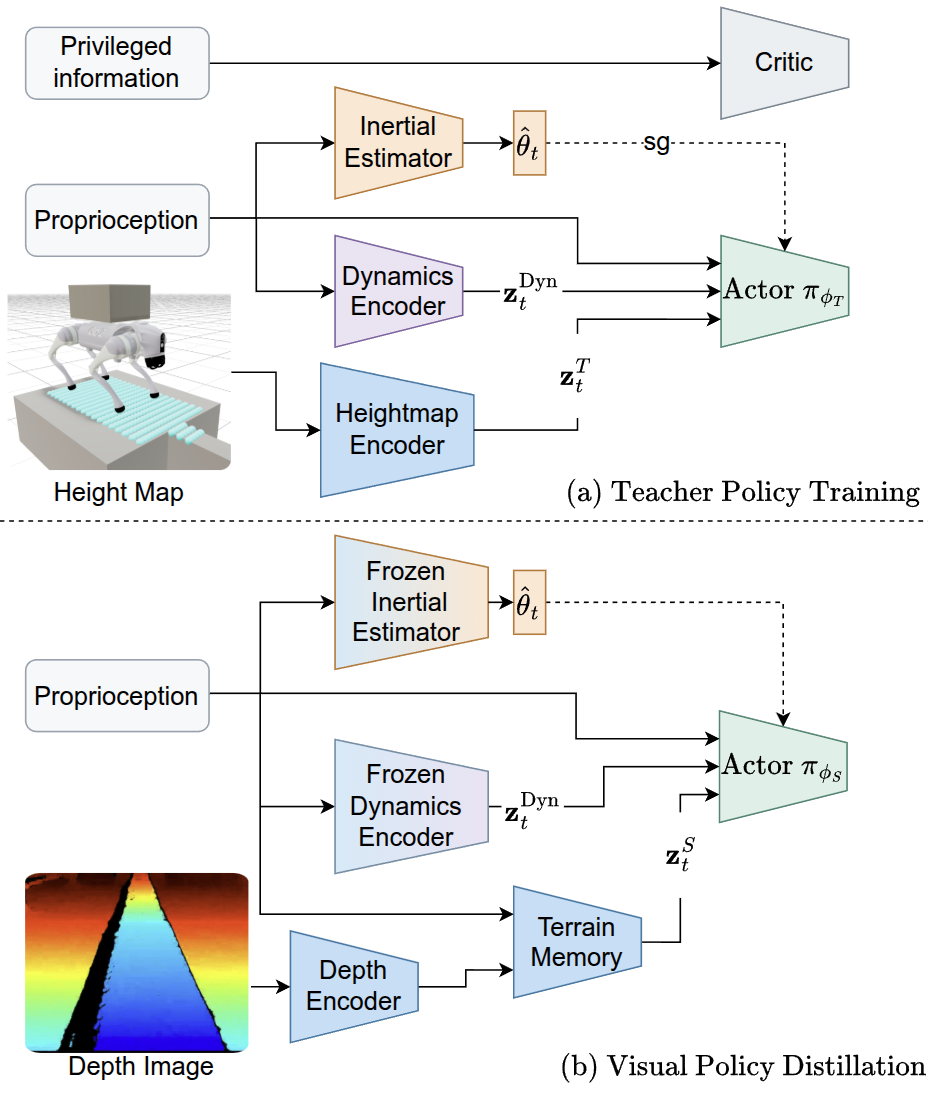}
    \caption{Overview of LOAD. sg denotes stop-gradient. }
    \label{fig:method}
\end{figure}

LOAD integrates dynamics perception and terrain perception for precision foothold locomotion under unknown payloads.
Two separate gated recurrent unit (GRU) networks process proprioceptive history to extract an implicit dynamics feature and estimate payload-induced changes in the torso's inertial properties relative to a nominal model.
A teacher policy uses terrain features encoded from privileged height maps and is distilled into a student policy that uses onboard depth images for deployment.
The locomotion policy combines the implicit dynamics feature, inertial estimates, and terrain features with current proprioceptive observations.
Fig.~\ref{fig:method} provides an overview of LOAD. 

\subsection{Observation and Action Spaces} \label{subsec:obs}
We train the teacher policy using proximal policy optimization (PPO)~\cite{schulman2017proximal} with an asymmetric actor--critic architecture.
The actor is a multilayer perceptron (MLP) policy $\pi_{\phi}$ with parameters $\phi$. The teacher $\pi_{\phi_T}$ and student $\pi_{\phi_S}$ have separate parameter sets and follow the same policy form:
\begin{equation}
    \mathbf{a}_t = \pi_{\phi}\!\left(
    o_t^{\mathrm{prop}}, \mathbf{p}_t, \mathbf{c}_t, \mathcal{N}(\hat{\boldsymbol{\theta}}_t), \mathbf{z}_t^\mathrm{Dyn}, \mathbf{z}_t^\mathrm{Ter}
    \right),
    \label{eq:actor}
\end{equation}
where $o_t^{\mathrm{prop}}$ is the current proprioceptive observation, $\mathbf{p}_t$ is the gait-phase encoding, $\mathbf{c}_t$ is the locomotion command, $\hat{\boldsymbol{\theta}}_t$ is the online estimate of the torso inertial parameter variation $\boldsymbol{\theta}$ defined in Sec.~\ref{subsec:dynamics_perception}, $\mathcal{N}$ normalizes the inertial coordinates, $\mathbf{z}_t^\mathrm{Dyn}$ and $\mathbf{z}_t^\mathrm{Ter}$ are the dynamics and terrain features, respectively. The policy action $\mathbf{a}_t$ is clipped, scaled, and offset to obtain joint-position targets, which are tracked by low-level proportional--derivative controllers.

The proprioceptive observation $o_t^{\mathrm{prop}}$ includes base angular velocity and projected gravity in the body frame, joint position offsets from the default configuration, joint velocities, and the previous action. The gait-phase encoding $\mathbf{p}_t$ contains the sine and cosine of the global gait phase with a period of 0.6 s and is set to zero for standing. The command $\mathbf{c}_t$ specifies the desired planar velocity and yaw rate.

During privileged training, the teacher receives a local height map $\mathbf{m}_t$. During deployment, this privileged observation is unavailable, and the student instead receives a depth image $\mathbf{d}_t$ from the onboard camera. Their encoded terrain features are denoted by $\mathbf{z}_t^T$ and $\mathbf{z}_t^S$, respectively, and are used as $\mathbf{z}_t^\mathrm{Ter}$ in Eq.~\eqref{eq:actor}. The terrain encoders are described in Sec.~\ref{subsec:terrain_perception}.

The MLP critic receives noise-free proprioceptive observations, commands, gait phases, ground-truth inertial parameters and height maps, together with base linear velocity and foot heights, air times, contact states, and forces.

% Because the current payload is not directly measured during deployment, an instantaneous observation does not fully determine the robot dynamics. We denote the proprioceptive history within the current episode by
% \begin{equation}
%     \mathcal{H}_t = \left\{o_s^{\mathrm{prop}}\right\}_{s=0}^{t}.
% \end{equation}
% Each observation already includes the previous action. The history $\mathcal{H}_t$ contains the robot responses needed to infer the unknown payload and is summarized recursively by the hidden state $\mathbf{h}_t$, as described in Sec.~\ref{subsec:dynamics_perception}.

\subsection{Teacher Policy Training} \label{subsec:policy}
% Unlike perceptive locomotion approaches that penalize foot placement relative to valid support regions~\cite{start2026}, we adapt locomotion rewards from Unitree RL Mjlab~\cite{unitree_rl_mjlab}, with a task-specific posture reference. The reference joint configuration is computed using leg inverse kinematics with a fixed target support width of 0.10~m chosen for beam traversal. This posture reward encourages a consistently narrow walking stance. Our focus is on precise foothold execution under unknown payloads rather than learning where footholds are admissible.

% The reward combines command tracking, posture and gait shaping, motion regularization, and a failure penalty. Table~\ref{tab:rewards} lists the reward terms and their weights $w_k$. At each control step, $r_t=\Delta t\sum_k w_k f_k(t), \Delta t=0.02~\mathrm{s}$, where $f_k$ denotes the unweighted reward term. The timestep factor applies to every term, including failure termination.

\block{Reward design}
We adapt the locomotion rewards from Unitree RL Mjlab~\cite{unitree_rl_mjlab} and introduce a posture reference tailored to narrow-beam traversal.
The reference joint configuration is computed using leg inverse kinematics with a reference support width of 0.10~m to encourage a consistently narrow walking stance.
Table~\ref{tab:rewards} lists the reward terms and their weights $w_k$.
At each control step, the total reward is $r_t=\Delta t\sum_k w_k f_k(t)$, where $f_k$ denotes the unweighted reward term and $\Delta t=0.02~\mathrm{s}$.
All terms, including the failure penalty, are scaled by $\Delta t$.

\begin{table}[t]
\centering
\caption{Rewards used for training teacher policy.}
\label{tab:rewards}
\fontsize{8}{9.5}\selectfont
\setlength{\tabcolsep}{3pt}
\renewcommand{\arraystretch}{1.08}
\begin{tabular}{@{}lr@{\hspace{8pt}}lr@{}}
\toprule
\textbf{Reward term} & \textbf{Weight} & \textbf{Reward term} & \textbf{Weight} \\
\midrule
Linear velocity tracking & $1$ & Soft joint limits & $-10$ \\
Angular velocity tracking & $1$ & Action rate & $-0.05$ \\
Gait contact agreement & $0.5$ & Joint acceleration & $-2.5\times10^{-7}$ \\
Support-width posture & $1$ & Foot clearance & $-1$ \\
Standing posture & $-1$ & Foot slip & $-0.25$ \\
Upright orientation & $-1$ & Touchdown impact & $-10^{-3}$ \\
Body angular velocity & $-0.05$ & Failure termination & $-200$ \\
Centroidal Ang. moment. & $-0.025$ & & \\
\bottomrule
\end{tabular}
\end{table}

Forward commands are sampled uniformly from $[0.5,0.7]$~m/s, with zero lateral and yaw commands; 5\% of environments receive standing commands. Hip/thigh/calf tolerances for the support-width posture reward are $(0.05,0.10,0.15)$~rad for standing and $(0.15,0.35,0.50)$~rad for moving commands. Gait-phase offsets are $(0,0.5,0.25,0.75)$ for $(\mathrm{FR},\mathrm{FL},\mathrm{RR},\mathrm{RL})$, with a stance duty factor of 0.8. The target terrain-relative foot clearance is 0.10~m, and soft joint limits span 90\% of the joint range. The action-rate term uses the clipped action before joint-target scaling and offset.

\block{Action regularization}
Clipping limits the executed actions but does not directly constrain the underlying policy means. To discourage excessive growth of the teacher action means, we adopt the action-bound regularization used in the DeepMimic implementation~\cite{peng2018deepmimic}, penalizing deviations beyond the clipping bounds:
\begin{equation}
\mathcal{L}_{\mathrm{bound}}
=\frac{1}{n_a}\mathbb{E}_t\!\left[
\sum_{i=1}^{n_a}\max\!\left(|\mu_{t,i}|-a_{\max},0\right)^2
\right],
\label{eq:action_bound_loss}
\end{equation}
where $\mu_{t,i}$ is the unclipped teacher action mean for joint $i$, $n_a=12$ is the action dimension, and $a_{\max}=3$ is the action clipping limit. This term is added directly to the PPO optimization loss with weight $\lambda_{\mathrm{bound}}=10^{-3}$, without timestep scaling.

\block{Termination}
Episodes terminate upon any of four failure conditions: body tilt exceeding $70^\circ$ from upright, non-foot terrain contact, self-collision, or a fall from the beam. Episodes also end when all four feet contact the terminal platform without failure (success), or at the 20~s time limit (timeout). Success and timeout alone incur no failure penalty.

\block{Terrain curriculum}
We use a terrain curriculum with 15 width levels. All environments begin on a 3.20~m-wide support surface. The nominal beam width decreases to 0.50~m, followed by 0.20~m, and subsequently decreases in 0.01~m increments to 0.08~m. Each environment advances one level after a successful crossing without failure and retreats one level after an unsuccessful episode, including a timeout, with the level bounded below by the initial level. After a successful crossing at the narrowest level, a new level is sampled uniformly from all levels. Beam lengths are sampled from 0.75--3.00~m, and beam widths beyond the initial level are perturbed uniformly by up to $\pm0.005$~m. Beam surface heights are sampled uniformly from 0.20--0.60~m.

\block{Domain randomization}
Table~\ref{tab:domain_randomization} summarizes the key dynamics and observation randomization settings. Proprioceptive noise bounds are ordered as angular velocity (rad/s), projected gravity, joint position (rad), and joint velocity (rad/s). 
At each episode reset, we uniformly sample an integer number of payload items from zero to eight and randomly stack them in the basket with support from below. Each item has a mass of 1~kg and dimensions of $0.15\times0.08\times0.03$~m. In simulation, the basket measures $0.30\times0.24\times0.15$~m (length $\times$ width $\times$ height). The stacking layout has $2\times3\times6$ available positions along these directions, with stacks allowed to extend above the rim. The combined base and payload mass, center of mass, and inertia are computed from their rigid-body composition, yielding coupled inertial variations.

\begin{table}[t]
\centering
\caption{Key domain randomization settings.}
\label{tab:domain_randomization}
% \fontsize{8}{11.5}\selectfont
\small
\setlength{\tabcolsep}{4pt}
\renewcommand{\arraystretch}{1.08}
\begin{tabular}{@{}lc@{}}
\toprule
\textbf{Parameter} & \textbf{Range / value} \\
\midrule
Payload item count & $\{0,\ldots,8\}$ \\
Foot friction coefficient & $[0.3,1.6]$ \\
Encoder bias (rad) & $[-0.02,0.02]$ \\
Action delay (ms) & $\{0,5,10\}$ \\
Proprio. noise bounds & $\pm(0.20,0.05,0.01,1.50)$ \\
Height map noise (m)$^{T}$ & $[-0.10,0.10]$ \\
Camera position offset (m)$^{S}$ & $[-0.01,0.01]$ \\
Camera angle bounds (rad)$^{S}$ & $\pm(0.02,0.10,0.02)$ \\
Depth multiplier$^{S}$ & $[0.98,1.02]$ \\
Depth Gaussian noise (m)$^{S}$ & $\sigma=0.01$ \\
Pixel / frame dropout prob.$^{S}$ & $0.002\,/\,0.05$ \\
Depth delay (ms)$^{S}$ & $\{0,20,40\}$ \\
\bottomrule
\end{tabular}
\par\vspace{3pt}
{\footnotesize
$T$/$S$: Teacher/Student only; unmarked rows apply to both. Intervals and discrete sets denote uniform sampling.\par}
\end{table}

% The reward encourages command tracking while penalizing foot placements outside the admissible support region: \begin{equation} \begin{split} r_t ={}& w_v r_v + w_{\omega} r_{\omega} + w_f r_{\mathrm{foot}} + w_b r_{\mathrm{balance}} \\ &- w_s p_{\mathrm{slip}} - w_{\tau} p_{\mathrm{torque}} - w_a p_{\mathrm{action}} - w_c p_{\mathrm{collision}} . \label{eq:reward} \end{split} \end{equation} Here, $r_v$ and $r_{\omega}$ reward linear-velocity and yaw-rate tracking. The precision foothold term is defined from the distance between each contacting foot and the valid support region: \begin{equation} r_{\mathrm{foot}} = \frac{1}{|\mathcal{C}_t|} \sum_{i\in\mathcal{C}_t} \exp\!\left( -\frac{d(\mathbf{p}^{\,i}_{f,t},\mathcal{S}_t)^2} {\sigma_f^2} \right), \label{eq:foothold_reward} \end{equation} where $\mathcal{C}_t$ is the set of contacting feet, $\mathbf{p}^{\,i}_{f,t}$ is the position of foot $i$, and $\mathcal{S}_t$ denotes the valid terrain support region. The balance reward regulates base orientation and lateral motion, while the remaining terms penalize foot slip, excessive torque, rapid action variation, and undesired collisions. Episodes terminate when the base or a non-foot body part contacts the ground, or when the robot leaves the traversable region.

\subsection{Online Dynamics Perception} \label{subsec:dynamics_perception} 

\block{Inertia Parameterization} We represent payload-induced changes through the inertial properties of the torso, including the rigidly attached basket and payload. Let $m$, $\mathbf{c}$, and $\mathbf{I}_C$ denote its mass, CoM position and inertia tensor about its CoM expressed in the body frame. We use a log-Cholesky parameterization~\cite{rucker2022smooth} to represent physically consistent torso inertial properties with unconstrained coordinates:
\begin{equation}
\frac{\tfrac{1}{2}\operatorname{tr}(\mathbf{I}_C)\mathbf{I}_3-\mathbf{I}_C}{m}
=\mathbf{U}\mathbf{U}^{\top},\qquad
\mathbf{U}=\begin{bmatrix}
e^{d_1} & s_{12} & s_{13}\\
0 & e^{d_2} & s_{23}\\
0 & 0 & e^{d_3}
\end{bmatrix}, \nonumber
% \label{eq:log_cholesky} 
\end{equation}
where $\mathbf{I}_3$ is the identity matrix. The corresponding ten coordinates are
\begin{equation}
\begin{split}
\mathcal{T}(m,\mathbf{c},\mathbf{I}_C)
=\bigl[&\tfrac{1}{2}\log m,\ d_1,d_2,d_3,s_{12},s_{23},s_{13},\ \mathbf{c}^{\top}\bigr]^{\top}. \nonumber
\end{split}
\end{equation}

\block{Dynamics Perception} The dynamics encoder and inertial estimator use separate GRUs, each followed by an MLP output head. Both the encoder and estimator receive the normalized proprioceptive observations, gait phases, and commands defined in Sec.~\ref{subsec:obs}. Omitting input normalization for brevity, their recurrent updates are
\begin{equation}
\mathbf{h}_t^{k}=f_{\mathrm{GRU}}^{k}\!\left(
o_t^{\mathrm{prop}},\mathbf{p}_t,\mathbf{c}_t,\mathbf{h}_{t-1}^{k}
\right),
\label{eq:gru}
\end{equation}
where $k\in\{\mathrm{imp},\mathrm{exp}\}$ denotes the implicit and explicit branches. Each branch has a 64-dimensional hidden state $\mathbf{h}_t^k$; neither parameters nor hidden states are shared between branches.

The dynamics encoder maps its hidden state to the implicit feature supplied to the actor:
\begin{equation}
\mathbf{z}_t^\mathrm{Dyn}=f_{\mathrm{MLP}}^{\mathrm{imp}}(\mathbf{h}_t^{\mathrm{imp}}),
\qquad \mathbf{z}_t^\mathrm{Dyn}\in\mathbb{R}^{6}.
\end{equation}
This branch is trained through PPO.

The estimation target for the inertial estimator is the coordinate difference from the nominal torso model:
\begin{equation}
\boldsymbol{\theta}
=\mathcal{T}(m,\mathbf{c},\mathbf{I}_C)
-\mathcal{T}(m_0,\mathbf{c}_0,\mathbf{I}_{C,0}),
\label{eq:inertial_parameterization}
\end{equation}
where subscript $0$ denotes the nominal torso model without the basket or payload. Thus, the explicit estimate represents inertial changes in log-Cholesky coordinates rather than the mass, center of mass, and inertia tensor components of the payload alone.
The inertial estimator maps its hidden state to the explicit estimate:
\begin{equation}
\hat{\boldsymbol{\theta}}_t=f_{\mathrm{MLP}}^{\mathrm{exp}}(\mathbf{h}_t^{\mathrm{exp}}),
\qquad \hat{\boldsymbol{\theta}}_t\in\mathbb{R}^{10}.
\label{eq:inertia_estimation}
\end{equation}
It is trained using simulated ground-truth inertial parameters with the supervised loss
\begin{equation}
\mathcal{L}_{\mathrm{id}}
=\mathbb{E}_t\!\left[
\frac{1}{10}\left\|
\mathcal{N}(\hat{\boldsymbol{\theta}}_t)-\mathcal{N}(\boldsymbol{\theta})
\right\|_2^2
\right],
\label{eq:id_loss}
\end{equation}
% \todoRen{Weighted loss may be better than normalized one.}
where $\mathcal{N}$ uses running means and standard deviations of the ground-truth coordinates collected during training. The same normalization is applied to predictions and targets. The estimator is trained alongside PPO using $\mathcal{L}_{\mathrm{id}}$ with unit weight, while its normalized output is detached before being passed to the actor.

The two branches provide complementary information: the explicit estimate supplies a representation supervised by physical inertial parameters, while the implicit feature retains task-relevant history without requiring all information to be expressed through the inertial coordinates. Both modules infer their outputs from proprioceptive history without visual input and are trained during teacher learning, then frozen during distillation.

\subsection{Terrain Perception} \label{subsec:terrain_perception}

Terrain perception encodes local support geometry into the terrain feature supplied to the locomotion policy.

\block{Height-map encoding}
The teacher receives a yaw-aligned height map $\mathbf{m}_t$ covering $0.80\times0.40$~m, centered 0.12~m forward of the torso origin. Longitudinal and lateral sampling intervals are 0.04~m and 0.02~m, respectively, yielding a $21\times21$ grid supplied at each policy step. An MLP-based height-map encoder compresses the map into the terrain feature $\mathbf{z}_t^{T}$.

\block{Depth-based encoding}
For deployment, the student instead uses a CNN to extract spatial features from the current $424\times240$ depth image $\mathbf{d}_t$ from a forward-facing camera. Depth is represented as distance along the camera optical axis, clipped to $[0.3,3.0]$~m, and normalized to $[0,1]$ before encoding. The terrain memory uses a GRU to integrate these features with normalized proprioceptive observations, gait phases, and commands to form the student terrain feature $\mathbf{z}_t^{S}$.
% \todoRen{Check whether architecture ablations are needed for the depth encoder and terrain memory.}
Proprioception provides motion cues, while the recurrent memory retains terrain information after the corresponding region leaves the camera's field of view.

\subsection{Policy Distillation} \label{subsec:policy_distillation}
We train the depth-based student to imitate the height-map teacher using DAgger-style online imitation learning~\cite{ross2011dagger}, which is common practice in perceptive locomotion~\cite{rudin2025parkour}.

\block{Imitation objective}
The student minimizes the action imitation loss $\mathcal{L}_{\mathrm{dist}}=\frac{1}{n_a}\mathbb{E}_t[\|\mathbf{a}_t^S-\bar{\mathbf{a}}_t^T\|_2^2]$, where $n_a$ is the action dimension, $\mathbf{a}_t^S$ is the deterministic action output of the student policy $\pi_{\phi_S}$, and $\bar{\mathbf{a}}_t^T$ is the clipped deterministic action target from the teacher policy $\pi_{\phi_T}$.
The action regularization in Sec.~\ref{subsec:policy} discourages but does not strictly prevent teacher action means from exceeding the clipping bounds. Imitating raw teacher actions could therefore cause the student to reproduce excessively large outputs, even though the excess beyond the clipping bounds is discarded before execution.

\block{Training procedure}
During rollout collection, the probability of executing teacher actions decreases linearly from one to zero, while the teacher provides supervision at every visited state.

Student training uses the same beam-width curriculum, physical randomization and proprioceptive noise as teacher training. To account for onboard depth sensing, we additionally apply the camera pose perturbations and depth corruption listed in Table~\ref{tab:domain_randomization}. Camera position bounds apply to each axis, and angle bounds follow roll, pitch, and yaw. Depth corruption includes spatially and temporally correlated missing regions and simulated stereo occlusions.

To retain the dynamics representations learned during teacher training, we freeze the teacher and the inherited dynamics encoder and inertial estimator, while training the student depth encoder, terrain memory, and actor. Jointly updating the two dynamics modules with the depth encoder and terrain memory led to distillation failure in our experiments.

\section{Experiments and Results}
\label{sec:experiments}

\subsection{Evaluation Protocol}

\block{Experiment Setup} We evaluate LOAD in simulation and on a Unitree Go2 equipped with a
torso-mounted shopping basket with a measured mass of 0.76~kg. The
payload consists of modular one-kilogram fitness-vest sandbags that can
be stacked or repositioned inside the basket. Reported payload masses
refer to the sandbags alone and exclude the basket. We vary the payload
configuration by changing the number and placement of the sandbags.
All variants use payload randomization during training, with payload masses
ranging from 0 to 8~kg.
Unless otherwise stated, tests use left-side payloads, a 3~m-long beam,
and a forward velocity command of 0.6~m/s. Unilateral loading introduces
a lateral CoM offset, increasing the demands on balance control.
We use traversal success rate as the primary evaluation metric.

\begin{table}[t]
\centering
\caption{Policy variants by implicit feature and inertial input.}
\label{tab:policy_variants}
\fontsize{8}{11.5}\selectfont
\setlength{\tabcolsep}{5pt}
\renewcommand{\arraystretch}{1.1}
\begin{tabular}{@{}lccc@{}}
\toprule
& \multicolumn{3}{c}{\textbf{Inertial input}} \\
\cmidrule(l){2-4}
\textbf{Implicit feature} & None & Estimated & Ground truth \\
\midrule
No  & Baseline & \policy{Explicit}     & \policy{Explicit}-GT \\
Yes & \policy{Implicit} & \textbf{LOAD} & LOAD-GT     \\
\bottomrule
\end{tabular}
\end{table}

\subsection{Teacher Policy Evaluation}

We compare the six policy variants summarized in Table~\ref{tab:policy_variants}.
All teacher policies are evaluated with the same height-map observations,
commands, and low-level controller.

\begin{figure}[t]
    \centering
    \includegraphics[width=\columnwidth]{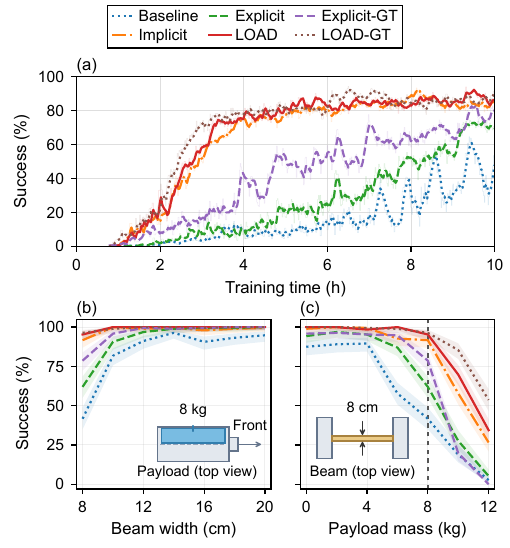}
    \caption{Teacher-policy comparison: (a) success rate over the latest 1,000 episodes on the narrowest beam (0.08~m). Faint lines show raw rates; bold lines show exponential moving averages with a 3~min time constant. (b) Varying beam width at 8~kg and (c) varying payload mass at 8~cm after 17,000 training iterations with left-side payloads. Shading in (b,c) shows pointwise 95\% Wilson intervals over 192 trials (64 per evaluation seed); the vertical dashed line in (c) marks the 8~kg training limit.}
    \label{fig:performance}
\end{figure}

In the reported training runs, policies with implicit dynamics
features reach high traversal success earlier than their counterparts
without these features (Fig.~\ref{fig:performance}(a)). This trend also
holds when ground-truth inertial parameters are available, as
illustrated by LOAD-GT versus \policy{Explicit}-GT.
Fig.~\ref{fig:performance}(b,c) compares the six teacher policies across
beam widths and payload masses. Three trends emerge.

First, the implicit dynamics feature improves performance across
different inertial inputs, as shown by Baseline versus \policy{Implicit}, \policy{Explicit}
versus LOAD, and \policy{Explicit}-GT versus LOAD-GT. Its
benefit persists even when ground-truth inertial parameters are
available.

Second, comparisons within the training payload range highlight
the value of inertial information and its accuracy. With an 8~kg payload on an 8~cm beam,
success rises from 41.7\% for Baseline to 62.0\% for \policy{Explicit} and
78.6\% for \policy{Explicit}-GT. However, the decline of \policy{Explicit}-GT on
heavier, out-of-distribution (OOD) payloads of 10--12~kg shows that ground-truth inertial
parameters alone do not guarantee generalization beyond the training
range.

Finally, on an 8~cm beam, LOAD achieves 95.3\% success with an
8~kg payload, compared with 91.7\% for \policy{Implicit}. At 10~kg,
the corresponding rates are 69.8\% and 57.8\%.
LOAD also achieves the overall success rate closest to
LOAD-GT: 96.5\% versus 97.8\%, averaged over all 49 tested
beam-width and payload-mass combinations, including OOD payloads.
This supports combining implicit
dynamics features with supervised inertial estimates.

\subsection{Identification Accuracy and Generalization}

We evaluate the LOAD teacher after 17,500 training iterations on a
20~cm-wide beam.
Physical domain randomization and proprioceptive noise are enabled.
The decoded physical quantities describe the combined torso,
basket, and payload, excluding the legs. CoM coordinates and inertia
components are reported relative to the base-link origin.

\begin{figure}[t]
    \centering
    \includegraphics[width=0.86\columnwidth]{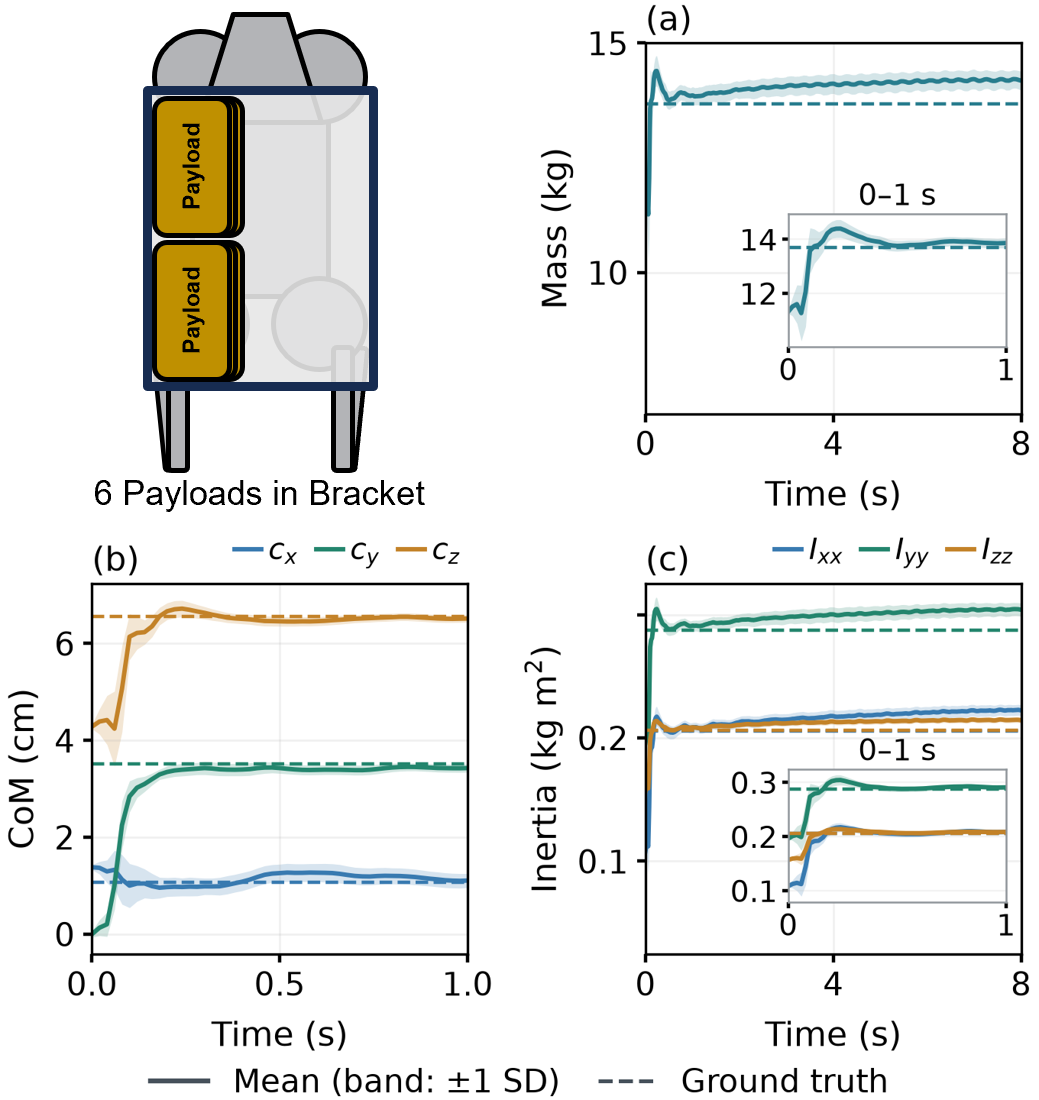}
    \caption{Identification with a 6~kg left-side payload over 1024 episodes. Solid curves and bands show the mean and $\pm1$ standard deviation; dashed lines indicate ground truth. The CoM panel and insets show the first second.}
    \label{fig:identification}
\end{figure}

\begin{figure}[t]
    \centering
    \includegraphics[width=\columnwidth]{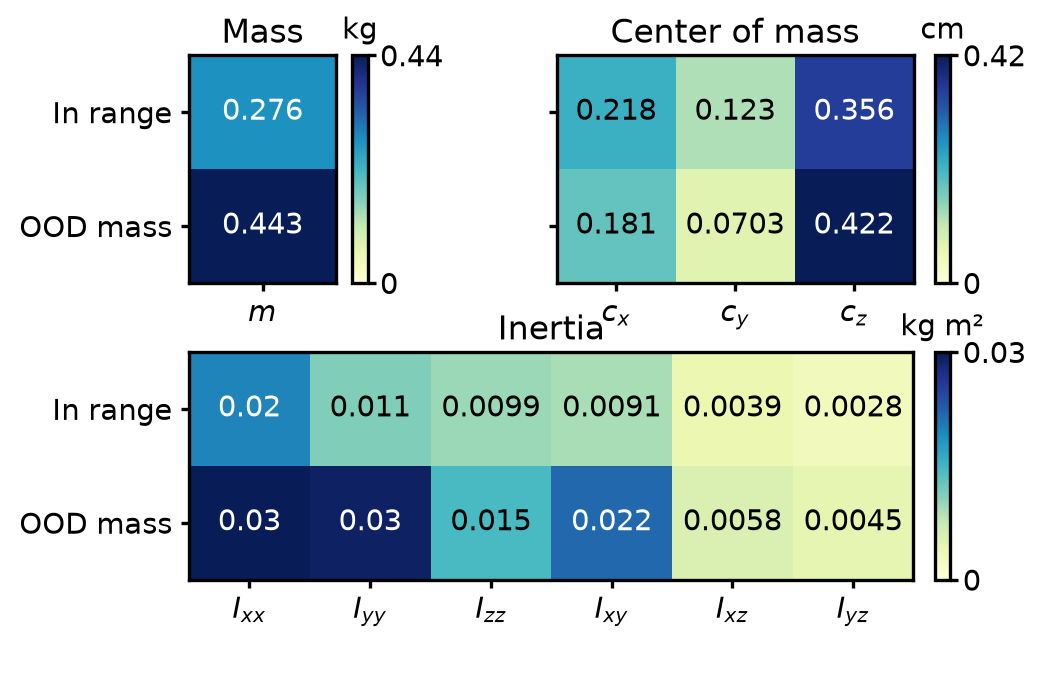}
    \caption{Full-episode mean absolute error (MAE) for random payloads: 0--8~kg within the training range and 9--12~kg outside it. Errors include the initial transient and give equal weight to each episode within a group.}
    \label{fig:identification_mae}
\end{figure}

Fig.~\ref{fig:identification} shows rapid recovery of the lateral
CoM offset, while mass and diagonal inertia estimates retain positive
biases during traversal. For aggregate evaluation, we sample 1,024 payload configurations with
integer masses uniformly distributed from 0 to 12~kg and use the
training stack-placement procedure. Fig.~\ref{fig:identification_mae}
shows higher mass, vertical CoM, and inertia errors for unseen heavier
payloads, while horizontal CoM errors remain small. This indicates
limited extrapolation of the inertial estimates beyond the training
mass range.

\subsection{Student Policy Comparison}
\label{subsec:distillation_performance}

\begin{figure*}[t]
    \centering
    \includegraphics[width=0.89\textwidth]{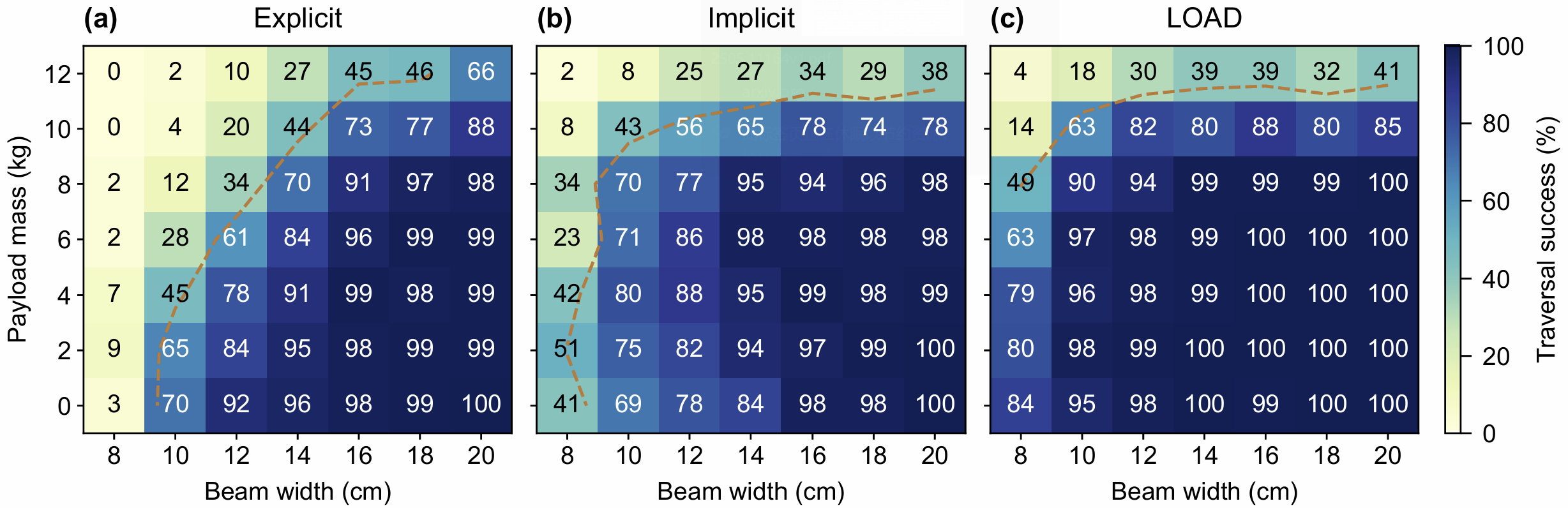}
    \caption{Post-distillation traversal success of depth-based student policies with left-side payloads: (a) \policy{Explicit}, (b) \policy{Implicit}, and (c) LOAD. Each cell reports success (\%) over 128 trials across four evaluation seeds. Dashed orange curves indicate interpolated 50\% success contours. }
    \label{fig:student_policy_performance}
\end{figure*}

To assess deployment performance, we compare the depth-based students
obtained by distilling \policy{Explicit}, \policy{Implicit}, and LOAD
(Fig.~\ref{fig:student_policy_performance}). We exclude Baseline due to
its low teacher-policy success rate, and \policy{Explicit}-GT and LOAD-GT
because they require ground-truth inertial parameters unavailable at
deployment.
The comparison evaluates
whether combining implicit dynamics features with explicit inertial
estimates remains beneficial when terrain observations come from depth
images. Physical domain randomization and proprioceptive noise are
enabled, while depth randomization is disabled during evaluation.
During distillation, we freeze the inherited dynamics encoder in
\policy{Implicit}, the inertial estimator in \policy{Explicit}, and both modules in
LOAD, as described in Sec.~\ref{subsec:dynamics_perception}.

LOAD retains a clear advantage on narrow beams within the training
mass range. At 8~kg on a 10~cm beam, its student achieves approximately
90\% success, compared with 70\% for \policy{Implicit} and 12\% for
\policy{Explicit}. On an 8~cm beam with a 6~kg payload, the corresponding
rates are 63\%, 23\%, and 2\%. These results support the complementary
roles of implicit and explicit dynamics information after visual
distillation. Generalization to heavier payloads remains limited:
at 12~kg, LOAD stays below 50\% success at every tested beam width,
and \policy{Explicit} performs better on the wider beams. Thus, LOAD's
advantage after distillation does not extend uniformly beyond the
training mass range.

% We perform ablations on the payload representation, training
% procedure, and perception inputs. For the payload representation, we
% compare mass-only conditioning, mass and first-moment conditioning,
% and the complete ten-dimensional inertial representation. The
% corresponding success rates are reported in
% Table~\ref{tab:ablation_payload}. The complete representation is
% expected to be most useful when the payload placement changes the
% rotational response and lateral body motion, whereas mass-only
% conditioning cannot distinguish payloads with the same total mass but
% different spatial distributions.

% Finally, we evaluate the effect of terrain perception by replacing the
% deployable depth student with the privileged height-map teacher. The
% teacher defines the terrain-perception upper reference, while the
% student measures the performance available from onboard sensing.
% This comparison allows the effect of terrain uncertainty to be
% separated from the effect of payload-dynamics uncertainty.

\subsection{Hardware Evaluation}

\begin{figure}[t]
    \centering
    \includegraphics[width=\columnwidth]{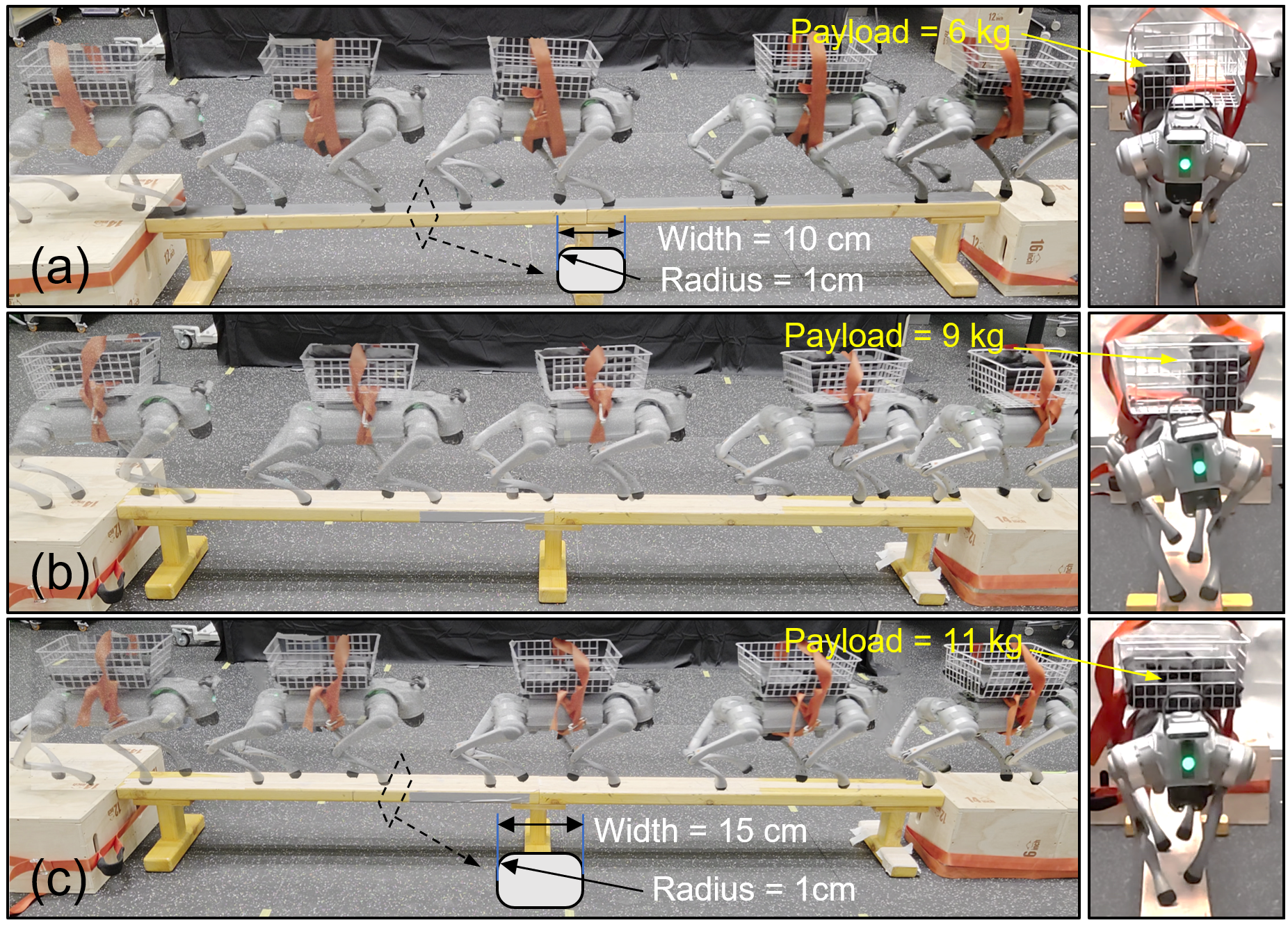}
    \caption{Hardware demonstrations of LOAD on Go2 robot: (a) a 6~kg right-side payload on a 10~cm beam, (b) a 9~kg left-side payload on a 15~cm beam, and (c) an evenly distributed 11~kg payload on a 15~cm beam. %Each row shows overlaid side-view snapshots (left) and a corresponding frontal view (right). 
    Note that the payload masses exclude the 0.76~kg basket, and the beams have round edges with $\sim$1cm radius which reduces actual support width.}
    \label{fig:hardware}
\end{figure}

\begin{figure}[t]
    \centering
    \includegraphics[width=\columnwidth]{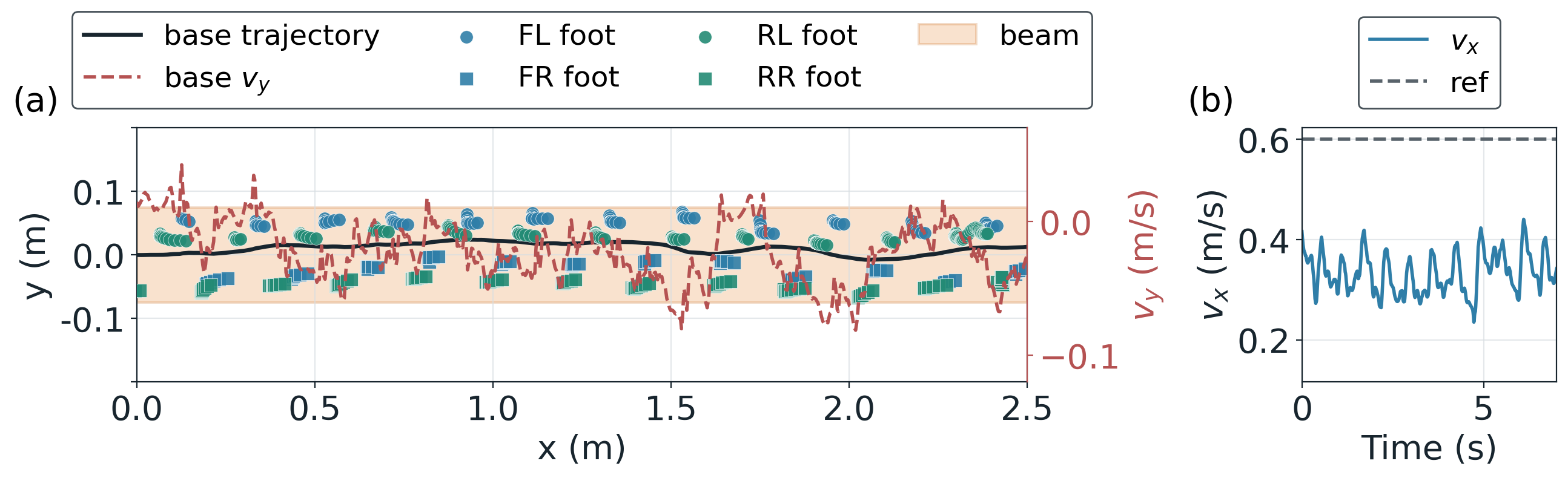}
    \caption{Beam traversal with a 6~kg right-side payload on a 0.15~m-wide beam: (a) foot placement, base trajectory, and estimated lateral velocity; (b) estimated and commanded forward velocities. Base and foot positions are reconstructed using a revised version of PRIME~\cite{prime2026} that operates without motion-capture input.}
    \label{fig:hardware_feet_velocity}
\end{figure}

We evaluate \policy{Implicit} and LOAD on hardware, as both perform
well in simulation and require no ground-truth inertial parameters.
Fig.~\ref{fig:hardware} shows representative LOAD traversals with
unilateral and evenly distributed payloads.

With payloads placed on one side, the maximum successfully tested
payloads on a $15\,\mathrm{cm}$ beam are $6\,\mathrm{kg}$ for \policy{Implicit} and
$9\,\mathrm{kg}$ for LOAD, with the latter placed on the left side
(Fig.~\ref{fig:hardware}(b)). On a $10\,\mathrm{cm}$ beam, the
maximum successfully tested payloads are $0\,\mathrm{kg}$ for
\policy{Implicit} and $6\,\mathrm{kg}$ for LOAD, with the latter
placed on the right side
(Fig.~\ref{fig:hardware}(a)). With an evenly distributed payload,
LOAD also traverses the $15\,\mathrm{cm}$ beam carrying $11\,\mathrm{kg}$
(Fig.~\ref{fig:hardware}(c)).
Including the 0.76\,kg basket, the combined basket and payload mass
reaches 11.76\,kg.
Unitree specifies maximum payloads of approximately 10--12\,kg across
Go2 variants~\cite{UnitreeGo2}, placing this narrow-beam
demonstration near the manufacturer's stated payload limits.

Fig.~\ref{fig:hardware_feet_velocity} shows lateral velocity oscillations
consistent with inverted pendulum dynamics during locomotion, while
the estimated forward velocity remains below the commanded value.

\subsection{Discussion}

Across the experiments, three consistent trends emerge. First, within the training mass range,
performance differences are small on wide beams and become pronounced
on narrow beams. Second, LOAD provides a task-level advantage over
\policy{Implicit} and approaches LOAD-GT performance for
in-distribution payloads. Third, OOD generalization remains limited
in both inertial estimation and control. Together, these results support the central
claim of LOAD: precise foothold locomotion benefits from perceiving
both the external terrain and the robot's current physical condition.

% \input{sections/05_results_and_analysis}

%%%%%%%%%%%%%%%%%%%%%%%%%%%%%%%%%%%%%%%%%%%%%%%%%%%%%%%%%%%%%%%%%%%%
 
\section{Conclusion and Future Work}
\label{sec:conclusion}

We presented LOAD, a locomotion approach that combines terrain
perception, explicit inertial estimation, and implicit dynamics
features for narrow-beam traversal under unknown payloads.
Separate recurrent modules recover physically interpretable inertial
changes and task-relevant history, while a privileged teacher using height maps
is distilled into a deployable student using onboard depth images. Comparisons across
six policy variants demonstrate the complementary value of explicit
and implicit dynamics information: history remains beneficial even
with ground-truth inertial parameters, and the LOAD teacher achieves
an overall success rate close to that of LOAD-GT over the tested conditions.
The benefit of the combined representation persists after visual distillation, particularly on
narrow beams within the training payload range. Hardware experiments
demonstrate traversal with unilateral payloads of 9\,kg on a 15\,cm
beam and 6\,kg on a 10\,cm beam.

Generalization beyond the training payload range remains limited.
Inertial estimates exhibit increased errors for heavier loads, while
performance degradation with ground-truth parameters indicates that
accurate identification alone does not ensure successful
extrapolation. Future work will investigate uncertainty-aware
conditioning and training strategies that improve both identification
and control under unseen dynamics. We will also extend the evaluation
to payload changes during locomotion and more varied support
geometries, examining how terrain and dynamics perception jointly
support reliable contact execution.

\balance
% \addtolength{\textheight}{-3.7cm}   % This command serves to balance the column lengths
%                                   % the textheight of the last page by a suitable amount.
%                                   % This command does not take effect until the next page
%                                   % so it should come on the page before the last. Make
%                                   % sure that you do not shorten the textheight too much.

%%%%%%%%%%%%%%%%%%%%%%%%%%%%%%%%%%%%%%%%%%%%%%%%%%%%%%%%%%%%%%%%%%%%%%%%%%%%%%%%

%%%%%%%%%%%%%%%%%%%%%%%%%%%%%%%%%%%%%%%%%%%%%%%%%%%%%%%%%%%%%%%%%%%%%%%%%%%%%%%%

%%%%%%%%%%%%%%%%%%%%%%%%%%%%%%%%%%%%%%%%%%%%%%%%%%%%%%%%%%%%%%%%%%%%%%%%%%%%%%%%
% \begin{comment}


%     \section*{ACKNOWLEDGMENT}
% The authors would like to thank Sequoyah Walters and Yifen Chen for inspirational dicussions, Heidi Fehr for advising us on Mocap system, and Peter Adamcyzk for letting us use his lab space for doing the experiments.

% \end{comment}

%%%%%%%%%%%%%%%%%%%%%%%%%%%%%%%%%%%%%%%%%%%%%%%%%%%%%%%%%%%%%%%%%%%%%%%%%%%%%%%%
%  \newpage
% \bibliographystyle{IEEEtran}
% \bibliography{reference}

% \section{Acknowledgement}
% We thank Runnan Chen for helping us build experiment apparatus. We also thank Jiarong Kang for helpful discussion on inertia identification.

\bibliographystyle{IEEEtran}
\bibliography{ref/gpt, ref/RL, ref/control, ref/estimation, ref/design_and_robot}
\end{document}